\documentclass[10pt, a4paper]{article}

\usepackage{lrec-coling2024} 
\usepackage{orcidlink}
\usepackage{multibib}
\newcites{languageresource}{Language Resources}
\usepackage{graphicx}
\usepackage{tabularx}
\usepackage{subcaption} 

\usepackage{soul}
\usepackage{enumerate}
\usepackage{enumitem}
\usepackage{caption}
\usepackage{titlesec}
\titleformat{\section}{\normalfont\large\bfseries\center}{\thesection.}{1em}{}
\titleformat{\subsection}{\normalfont\bfseries\raggedright}{\thesubsection.}{1em}{}
\titleformat{\subsubsection}{\normalfont\normalsize\bfseries\raggedright}{\thesubsubsection.}{1em}{}
\renewcommand\thesection{\arabic{section}}
\renewcommand\thesubsection{\thesection.\arabic{subsection}}
\renewcommand\thesubsubsection{\thesubsection.\arabic{subsubsection}}
\usepackage{lipsum}
\usepackage{epstopdf}
\usepackage[utf8]{inputenc}

\usepackage{hyperref}
\usepackage{xstring}

\usepackage{color}

\title{3D-LEX v1.0\\3D \textbf{Lex}icons for American Sign Language \\and Sign Language of the Netherlands}
\name{O. Ranum{$^1$} \orcidlink{0000-0001-8627-6259}, G. Otterspeer{$^1$} \orcidlink{0009-0007-5289-9379}, J.I. Andersen{$^1$} \orcidlink{0009-0000-4996-7421}\vspace{1mm},\\ {\bf \large R.G. Belleman{$^2$} \orcidlink{0000-0001-6576-8350}, F. Roelofsen{$^1$}\orcidlink{0000-0002-3224-5251}}} 

\address{\\ University of Amsterdam \\
         {$^1$}: Institute for Logic, Language and Computation, University of Amsterdam \\
         {$^2$}: Computational Science Lab, Informatics Institute, University of Amsterdam \\
         oline.ranum@student.uva.nl,\{g.otterspeer, j.andersen, r.g.belleman, f.roelofsen\}@uva.nl\\}

\abstract{
In this work, we present an efficient approach for capturing sign language in 3D, introduce the 3D-LEX v1.0 dataset, and detail a method for semi-automatic annotation of phonetic properties. Our procedure integrates three motion capture techniques encompassing high-resolution 3D poses, 3D handshapes, and depth-aware facial features, and attains an average sampling rate of one sign every 10 seconds. This includes the time for presenting a sign example, performing and recording the sign, and archiving the capture. The 3D-LEX dataset includes 1,000  signs from American Sign Language and an additional 1,000 signs from the Sign Language of the Netherlands. We showcase the dataset utility by presenting a simple method for generating handshape annotations directly from 3D-LEX. We produce handshape labels for 1,000 signs from American Sign Language and evaluate the labels in a sign recognition task. The labels enhance gloss recognition accuracy by \texttt{5}\% over using no handshape annotations, and by \texttt{1}\%  over expert annotations. Our motion capture data supports in-depth analysis of sign features and facilitates the generation of 2D projections from any viewpoint. The 3D-LEX collection has been aligned with existing sign language benchmarks and linguistic resources, to support studies in 3D-aware sign language processing.
\\
 \newline  }

\begin{document}

\maketitleabstract



\section{Introduction}

Sign language processing (SLP) is a dynamic research area concerned with advancing computational methods for sign languages (SL). This multidisciplinary field encompasses tasks such as the automatic understanding, recognition, translation and production of sign language, contributing to a more inclusive future in language technology.

Despite receiving increased attention across computer sciences (\citealp{koller2020, Rastgoo_2021}),  SLP remains less developed compared to other areas within Natural Language Processing \cite{yin2021}. A significant factor contributing to this disparity is the lack of large-scale, high-quality, and publicly accessible sign language corpora \cite{bragg2019}. Notably, the majority of these datasets are recorded with cameras that view signers from a single, (near-)frontal perspective \cite{sk2022}. This scarcity of data impedes modern machine-learning algorithms from learning robust sign representations grounded in the three-dimensional nature of sign languages.

\begin{figure}[!b]
    \centering
    \setlength{\fboxsep}{0pt} 
    \fbox{\includegraphics[width=0.89\columnwidth]{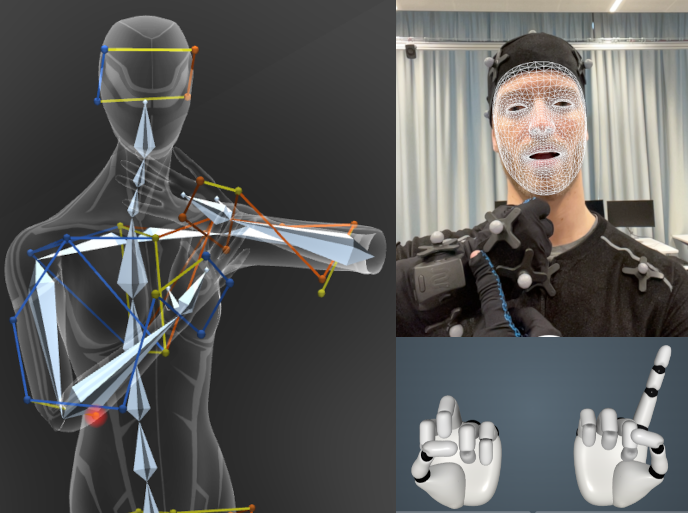}} 
    \caption{\textbf{Motion capture techniques}: The NGT sign \textit{'mango'} captured with the three collection techniques. Left:  Pose data captured with Vicon Motion Capture displayed in Shogun Live; Top right: face features captured with Live Link Face (Epic Games); Bottom right: handshapes captured with gloves displayed in Hand Engine (StretchSense).}
    \label{fig:intro}
\end{figure} 

Literature supports that dept-awareness and viewing angle matters in both human \cite{Watkins2024} and machine \citelanguageresource{Liquing22, RASTGOO2020113336} SL understanding. This implies that representations should reflect a degree of 3D awareness, or risk reduced accuracy under normal real-world conditions, such as non-frontal viewpoints. 

While systems such as OpenPose \cite{Cao2016} enable the estimation of 3D poses from video footage, the precision of such reconstructions is in principle lower than the accuracy achieved through direct 3D motion capture techniques \cite{jedlicka2020}. Navigating imperfectly reconstructed 3D representations can pose significant challenges for downstream SLP tasks.

Providing a 3D ground truth to existing datasets could significantly improve the feasibility of many SLP tasks. Against this backdrop, we introduce 3D Lexicons (3D-LEX) for American Sign Language (ASL) and the Sign Language of the Netherlands (Nederlandse Gebarentaal, NGT). The 3D-LEX datasets include 1,000 isolated signs from each language recorded with three distinct motion capture techniques, as illustrated in Figure \ref{fig:intro}. The vocabularies have been aligned with existing SL resources, including the WLASL \cite{Dongxu2020} and SEMLEX \cite{Kezar_2023} benchmarks for isolated sign recognition, the ASL-LEX 2.0 \cite{Sehyr2021} lexicon and the SignBank NGT (SB NGT) lexicon \citelanguageresource{Crasborn2020NGT}. The 3D-LEX dataset facilitates the generation of 2D projections from any viewpoint and supports in-depth analysis of sign language features, offering several key advantages:

\vspace{-0.3cm}
\paragraph{Automatic recognition of phonetic properties:} High-resolution 3D data allows for detailed studies of sign language features, including handshapes, place of articulation, and orientations.

\vspace{-0.3cm}
\paragraph{Multi-view SL recognition:} Ground truth 3D representations facilitate the rendering of synthetic multi-view 2D data from any angle and translation. This data can be used to train models that are capable of \textit{multi-view SL recognition}, a task that has received little attention in the SLP literature so far.

\vspace{-0.3cm}
\paragraph{SL production for XR applications:} 
Current work on SL production focusing on 2D outputs, such as synthetic photorealistic videos or 2D skeleton animations, are not directly suitable for Extended Reality (XR) applications. While reconstructing 3D motion from multiple 2D views is an area of active research, leveraging 3D data to produce 3D animations currently still offers a more effective and accurate approach.

\paragraph{}
\vspace{-0.3cm}

The 3D-LEX v1.0 dataset was developed during our initial exploration of motion capture equipment for capturing three-dimensional sign representations. We acknowledge that the methodology outlined in Section 3 presents significant opportunities for improvement. Specifically, ensuring consistency in data quality will be a primary objective in our future efforts. Nevertheless, even in this nascent stage of development, we could demonstrate the utility of the 3D-LEX data. In Section 4 we showcase how the dataset can be leveraged to produce semi-automatic annotations of handshapes. Evaluating the annotations in a downstream isolated sign recognition (ISR) task demonstrates that the labels achieved parallel benefits to leveraging annotations provided by linguists. We discuss several observed limitations and prospects for improvement in Section 5, and Section 6 highlights some ethical considerations.

\section{Background}\label{sec:bcg}

\subsection{Sign Language}

Sign languages are visual, complete, and natural languages, each with a distinct structure, grammar, and lexicon. They employ a combination of manual markers (\textit{e.g.} handshapes, hand location, palm orientation and movements) and non-manual markers (\textit{e.g.} mouthings, facial expressions, gaze) to convey meaning \cite{Stokoe2005}.  Sign languages serve as the primary language in Deaf communities.

\subsection{Sign Language Datasets}
The majority of publicly available resources demonstrating sign language are captured in video. These datasets consists of either isolated signs (\textit{e.g.} \citealplanguageresource{Sehyr2021, Athitsos2008, Kezar_2023, Jose_2018, Dongxu2020}) or continuous sign sentences (\textit{e.g.}
\citealplanguageresource{Argis_2010, sCHEMBRI_2013}). Key distinguishing features between the collections include the source language, signer variability, data scope, linguistic domain, and the availability and quality of annotations. 
 
 Most datasets comprise RGB video formats, but they may also include depth estimations or skeletal poses generated from joint approximations.  While these datasets usually feature a single, (near-)frontal viewpoint, there is a growing trend in lab-curated datasets to provide multiple viewing angles  (\textit{e.g.} \citealplanguageresource{Duarte_2020, Mopidevi2023, RASTGOO2020113336, Liquing22}).  Depth cameras have been used to capture 3D positioning, for example using the Kinect depth sensor  (\textit{e.g.} \citealplanguageresource{Oszust2013,cooper2012,Huang_2018}). For an extensive summary of sign language datasets, refer to \citetlanguageresource{kopf_maria_2021}. 

 Datasets facilitating 3D awareness in sign representations either leverage 
 depth estimations or 3D reconstruction techniques. For the creation of more precise 3D representations, numerous motion capture datasets have been curated  (\textit{e.g.} \citealplanguageresource{luhuenerfauth2010,Alexis2006,Benchiheub}), typically to generate signing avatars \cite{bragg2019} or for exploring automatic synthesis of sign language utterances (\textit{e.g.} \citealplanguageresource{jedlicka2020,Gibet2018}). 
 
\section{The 3D-LEX Dataset} \label{sec:method}

\subsection{Data Scope}
The 3D-LEX v1.0 dataset includes lexical datasets sampled from ASL and NGT, where the scope was defined to ensure integration with existing benchmarks. A total of 1,000 signs are selected from each language, and recorded with two data collection techniques to capture manual markers and one technique to capture non-manual markers\footnote{The data is available under a \href{https://creativecommons.org/licenses/by/4.0/}{CC BY 4.0 license} at \href{https://osf.io/g7u9c/?view_only=8090319e12aa4fd991d81e369a1cbd88}{osf.io/g7u9c/?view\_only=8090319e12aa4fd991d81e369\\a1cbd88}}. We release three distinct data formats corresponding to the different capturing techniques, and one component integrating handshapes and body pose data.

\vspace{-0.3cm}
\paragraph{Handshape Data} The handshape(s) of each sign is captured with the StretchSense Pro Fidelity Motion Capture Gloves\footnote{\href{https://stretchsense.com/mocap-pro-fidelity-glove-2/}{stretchsense.com/mocap-pro-fidelity-glove-2/}}. The gloves measure the splay and bend of the fingers, alongside the relative rotation of each joint within the hand. The available data include the stretch sensor readings and exported FBX\footnote{A 3D model file facilitating the transfer of animation data between various modeling applications including Maya, Blender, and Unreal Engine.} files. Detailed guidance on interpreting and assessing StretchSense data can be found in the project's Git repository for data evaluation\footnote{\href{https://github.com/OlineRanum/SAPA}{github.com/OlineRanum/SAPA}}. 

\vspace{-0.3cm}
\paragraph{Body Pose Data} The place of articulation, movement, and body pose of each sign is captured using a Vicon (V) Motion Capture setup with optical markers. The raw marker location data is published, alongside processed FBX data, which has been exported via Shogun Post.

\vspace{-0.3cm}
\paragraph{Face Blendshape Data} Facial features are captured as blendshapes with the Live Link Face\footnote{\href{https://apps.apple.com/us/app/live-link-face/id1495370836}{apps.apple.com/us/app/live-link-face/id1495370836}} (LLF) application and ARKit on iPhone. 

\vspace{-0.3cm}
\paragraph{Retargeted Animation Data} For sign language production and animation purposes, we release FBX files containing the body pose data and the handshapes.

\subsection{Production Method}
To efficiently capture the lexicons, we have developed a recording pipeline that achieves an average capture time of 10 seconds per sign. This includes the time for sign demonstration, performance, recording, and storage of the captured sign, though it varies with the sign's length. Setup preparations, which involve fitting the suit, positioning markers, and calibrations, require approximately 1 hour with our current method.

\subsubsection{Recording Setup and Procedure}
Our studio setup includes a designated detection zone for the Vicon cameras, an iPhone equipped with Live Link Face mounted on a tripod, one screen to display glosses and reference videos, and a second screen to showcase the recordings for immediate evaluation. 

A triple-foot pedal system facilitates the remote operation of the motion capture control system. Each pedal is configured for a distinct function: The left pedal triggers the start and stop of recordings across all three motion capture systems simultaneously; the middle pedal stores the latest recording and issues a request to the SignCollect platform to display the next gloss in the vocabulary; and the right pedal is used to proceed to the next sign without saving any data. Signcollect is a platform developed to enable the efficient processing of glosses, providing a studio interface managed by gesture or pedal control. For details on the SignCollect platform consult \citet{Otterspeer2024}.

The capture process for a single sign involves the following steps: First, the signer assumes an upright posture, with arms relaxed at their sides in a neutral position. By pressing the right pedal, a sign is prompted from the SignCollect platform, and the sign's gloss and a reference video are displayed on one of the screens. A recording is started by pressing the left pedal, and the signer performs the sign and returns to the neutral stance before the recording is ended with another press of the left pedal. The recorded data is automatically exported to SignCollect and visualized on an avatar rendered with Unreal Engine v5.3, allowing the signer to immediately review the quality of the data. If the data's quality is satisfactory, the signer can advance to the next gloss by pressing the right pedal, which saves the preceding recording. Should the sign's execution be deemed inadequate, the signer can repeat the recording by pressing the left pedal again or proceed by pressing the right pedal.  For visualizing the sign we created an avatar in \textit{Ready Player Me Studio}, a cross-platform avatar generator that allows users to build avatars for general purposes.

A total of five signers contributed to capturing the ASL and NGT vocabularies. The signers were given two options to operate the pedal. Either they could control the pedal and capture process themselves, or they could delegate the pedal control to a team member. Preferences varied, with three signers opting for controlling the pedal themselves and two preferring assistance to concentrate on signing. Details regarding the number of words recorded by each signer per language and pedal control preferences are provided in table \ref{tab:nl}.

The control system and comprehensive details about the pipeline are available on GitHub\footnote{\href{https://github.com/OlineRanum/GLEX_Controller}{github.com/OlineRanum/GLEX\_Controller}}. In the following paragraphs we describe each motion capture component in greater detail. 

\subsubsection{Vicon Motion Capture System}

\begin{figure}[!t]
    \centering
    \setlength{\fboxsep}{0pt} 
\fbox{\includegraphics[height = 4cm]{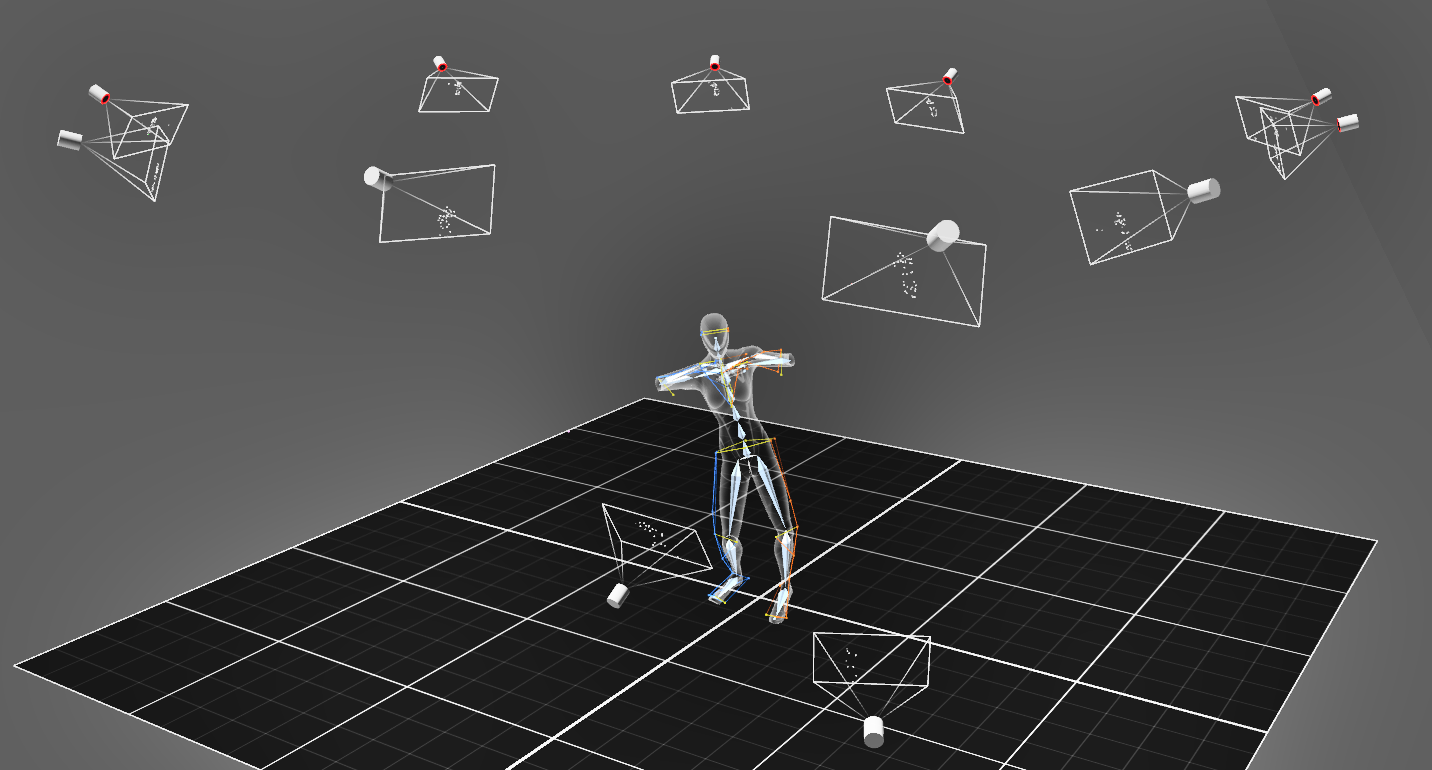}} 
    \caption{\textbf{Setup of the Vicon detection zone}: The illustration indicates the placement of the Vero Cameras on the rig and in front of the signer.}
    \label{fig:layout}
\end{figure}

\textit{Setup:} 
A Vicon rig is affixed to the ceiling, equipped with ten Vicon Vero v2.2 optical motion capture cameras\footnote{\href{https://www.vicon.com/hardware/cameras/vero/}{vicon.com/hardware/cameras/vero/}}, as detailed in Figure \ref{fig:layout}. To mitigate occlusions, particularly those caused by the lower hands of the signer, an additional two Vicon Vero cameras are placed on the floor in front of the signer.

\begin{figure}[!t]
    \centering
    \setlength{\fboxsep}{0pt}  \fbox{\includegraphics[height = 4cm]{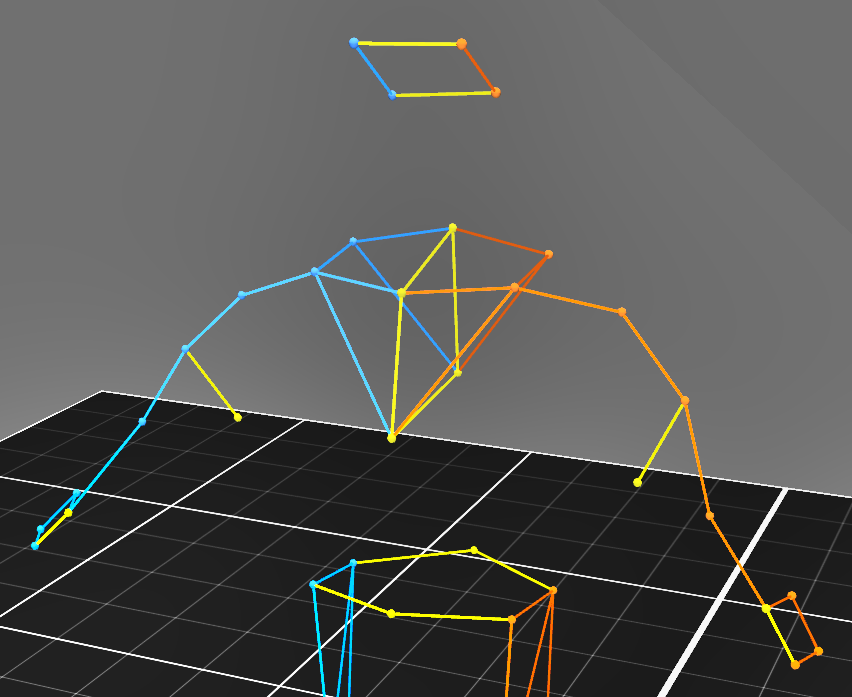}}
    \caption{\textbf{Marker layout for the Vicon system}: Layout according to FrontWaist 53-marker set template, displayed on signer in Shogun Live.}
\label{fig:marker_layout}
\end{figure}

The markers are placed on the signer following the standard Vicon FrontWaist 53-marker set template, as displayed in Figure \ref{fig:marker_layout}. Shogun Post is used to make a retarget for the motion capture data, which is used during recording to stream the data to Unreal Engine from Shogun Live.

\textit{Calibration:} 
For calibrating the Vicon camera system, we adhere to the built-in calibration protocol provided by Vicon. To ensure consistency in the calibration and that the origin remains approximately in the same position across multiple recording sessions, we place masking tape on the floor. This tape serves a dual purpose: one set of markings indicates the precise location for positioning the calibration wand during each calibration process. Another set of tape strips marks the designated spot where the signer is to stand during recordings. 

\textit{Software Specifications:}
To manage the Vicon camera system, we utilize Shogun Live 1.11, and to perform the retarget of the motion capture data we use Shogun Post 1.11.

\subsubsection{StretchSense Gloves}

\textit{Setup:} The StretchSense Pro Fidelity gloves interface with Hand Engine Pro through two USB dongles, which are docked on a separate Dell Universal Dock (UD22) to ensure adequate power supply. Hand Engine is configured to receive remote triggering from Shogun Live, and to retarget animation data directly to Unreal Engine.

\textit{Calibration:}
The StretchSense Pro Fidelity gloves are calibrated using the calibration functionality of the Hand Engine software, which involves capturing pre-defined hand poses, to match the recorded output to an individual's hand. Our procedure combines general-purpose poses with specialized ones to customize the glove's fit for each user to capture sign language. 

\begin{enumerate}[label=\roman*.]
    \item \textbf{Express Calibration Poses:} Our general-purpose hand pose set corresponds to the express calibration poses available in Hand Engine, which comprises five common handshapes. 
    \item \textbf{Advanced Calibration Poses:} A more detailed hand pose library was developed, incorporating the most commonly occurring handshapes found in the 3D-LEX NGT (20 poses) and ASL (25 poses) vocabulary, as labeled by linguists in the aligned resources. The advanced pose libraries have been made accessible on GitHub. 
\end{enumerate}

We employ the training functionality of Hand Engine to fit the gloves' output data specifically to the signer. We configure all calibration poses to the blend pose mode, a Hand Engine feature that uses the calibration poses as landmarks in a continuous motion space, and interpolates between these poses to yield continuous outputs. The gloves are calibrated and retrained each time a signer puts them on to maintain accuracy. 

Following initial consultations with StretchSense about employing the Pro Fidelity gloves for sign language capture, we developed the specific number of poses and this calibration scheme. However, throughout the creation of 3D-LEX and subsequent discussions, it became evident that the calibration scheme was not ideal. We acknowledge this shortcoming and will reevaluate the calibration process in future works. For a discussion of these limitations and suggestions for potential improvements, please see Section 5.

\textit{Software Specifications:} The StretchSense Pro Fidelity gloves are operated with the Hand Engine Pro software, version 3.0.6.

\subsubsection{Live Link Face}
\textit{Setup:} An iPhone is mounted on a tripod, which is placed directly in front of the signer. Recordings are started, stopped, and saved automatically by the remote triggers. 

\textit{Calibration:} Live Link Face was not calibrated per signer. However, this functionality is available in the Live Link Face application and should be explored in a later version of the dataset.

\textit{Hard- and software Specifications:} We use an iPhone 13 Pro and run Live Link Face version 1.3.2 with iPhone AR Kit.


\subsection{Dataset Characteristics}

The recording procedure introduces several recurring patterns into the raw data. Notably,  the initial and final arm and hand positions often adopt a neutral stance, with the handshape closely resembling a `5' handshape (refer to Figure \ref{fig:failure_mode}). This results in, for instance, parts of the handshape recordings capturing signals that are not characteristic of a particular sign. This includes handshapes observed during the transition from a neutral state to the sign's active posture, or when a sign involves a series of distinct handshapes, resulting in recordings that capture multiple pose signals within a single sign. An illustration of a typical temporal series according to the Euclidean distance is provided in Figure \ref{fig:ls}.

Data captured using LLF presents a non-uniform sampling rate, as frames are only recorded upon detected changes in the current state of the sensor. Conversely, the body poses captured with the Vicon system and handshapes captured with StretchSense are sampled uniformly.

\begin{figure}[!h]
    \centering
    \includegraphics[width = \columnwidth]{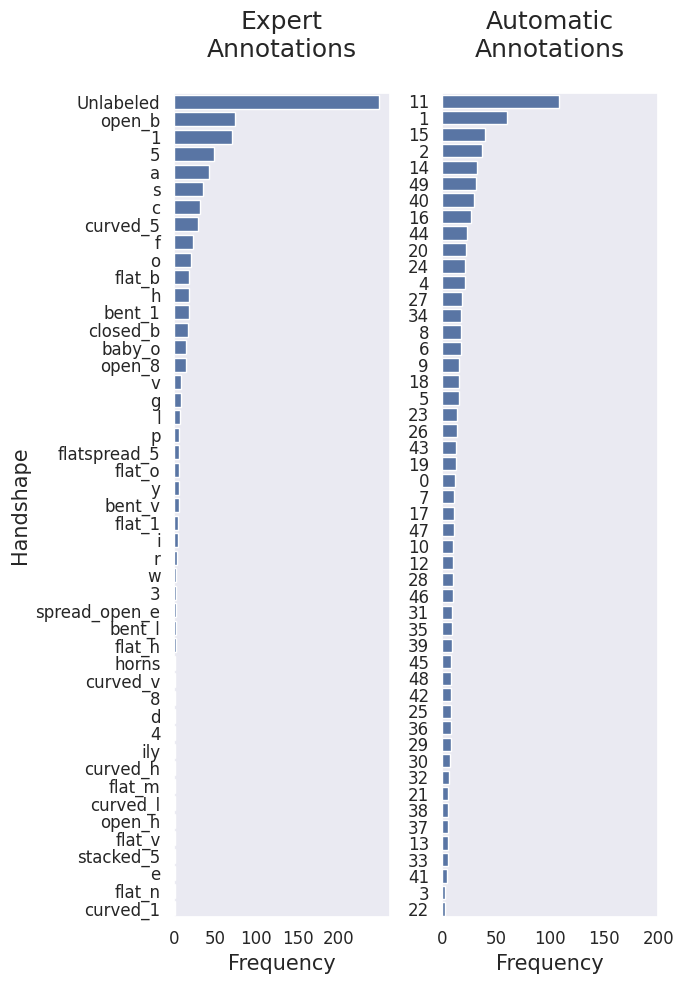}
    \caption{\textbf{Distributions of handshapes in the 3D-LEX vocabulary}: the distribution of handshapes as identified by (a) human experts and (b) the automated annotation process described in Section 4.1. The automatic annotations assign arbitrary cluster IDs to different groups of handshapes determined through a K-means clustering method. It's important to note that these handshape cluster IDs may not directly correspond to the linguistic labels used by human experts in Subfigure 4.a.}
    \label{fig:dist-handshapes}
\end{figure}

The lexicons include a variety of handshapes. Figure \ref{fig:dist-handshapes}.a showcases the distribution of handshapes in the ASL Lexicon, annotated by sign language linguists in the ASL-LEX resource.  

\paragraph{Signer characteristics}
All participants are native signers, who acquired sign language from an early age. Details about each signer's primary language, along with their preferences for operating the pedal, are provided in Table \ref{tab:nl}.

\begin{table}[!b]
\begin{center}
\begin{tabularx}{\columnwidth}{|X|l|l|l|l|l|}
      \hline
        \textbf{Signer ID} & 01   & 02  & 03  & 04  & 05 \\ \hline
        \textbf{Native} &&&&&\\ 
        \textbf{Language}&NGT&NGT&NGT&NGT&ASL\\\hline
        \textbf{NGT }&&&&&\\ 
        \textbf{Signs}&10&400&590&0&0\\ \hline
        \textbf{ASL}&&&&&\\
        \textbf{Signs}&155&12&0&644&189\\ \hline
        \textbf{Pedal}&&&&&\\
        \textbf{Control}&YES&YES&YES&NO&NO\\ \hline
\end{tabularx}
\caption{\label{tab:nl}\textbf{Signer Characteristics}: Native background of each signer and preference for operating (YES) or delegating (NO) the control of the pedal.}
 \end{center}
\end{table}

\paragraph{Alignment with existing SL resources}
The vocabularies of 3D-LEX have been aligned with existing SL resources to promote research integrating 3D data with datasets comprised of video data and linguistic databases. Table \ref{tab:benchmarks} lists the number of glosses in 3D-LEX overlapping with the vocabularies of the aligned resources, the number of sign pose estimations from example videos available for the glosses in the datasets, and the number of glosses that have been provided with expert human annotations for the dominant hand.

The 3D-LEX ASL vocabulary was selected to ensure that a minimum of five reference videos per sign are available in each ASL dataset. Currently, no dataset with multiple reference videos per gloss exists for NGT, but we anticipate that this situation will change in the future. Currently, the SB NGT lexicon  \citelanguageresource{Klomp2024, Crasborn2020NGT} provides one reference video for each gloss in the 3D-LEX NGT vocabulary.

\begin{table}[!h]
\begin{center}
\begin{tabularx}{\columnwidth}{|l|X|X|X|}
      \hline
        &\textbf{SEMLEX}&\textbf{WLASL}& \textbf{SB NGT}\\ \hline
         & ASL & ASL & NGT \\\hline
        Vocabulary & 1,000& 1,000 & 1,000\\\hline
        Reference & & & \\
        Videos & 49,274&12,051 & 1,000\\
      \hline
        Expert HS &921& 695 & 888 \\ \hline
\end{tabularx}
\caption{\label{tab:benchmarks}\textbf{Alignment with other datasets}:  The vocabulary overlap, the number of available reference videos, and the number of available expert handshape annotations for the 3D-LEX vocabulary in the SEMLEX, WLASL, and SB NGT datasets.}
 \end{center}
\end{table}


\section{Evaluation}\label{sec:eval}

To demonstrate the utility of 3D sign data we turn to one of the envisioned benefits mentioned initially: the facilitation of automatic phonetic labeling. In particular, we present a baseline method for semi-automatic handshape annotation. The efficacy of the annotations is evaluated in an ISR task, through comparison with labels provided by linguists and against scenarios devoid of any labels.

While we expect that the data can be used to label other phonetic properties (\textit{e.g.}\ hand location, movement, orientations, eyebrow position) we here zoom in on the handshapes. This is an intentional choice, as we consider the use of StretchSense gloves to be the most experimental data acquisition technique for sign language capture. The development of semi-automatic annotation methods benefits both linguistic research and various SLP tasks, including recognition and production.

\subsection{Semi-Automatic Handshape Annotations}
In this section, we demonstrate one simple approach for generating phonetic annotations derived from the 3D-LEX handshape data. Due to the absence of an NGT benchmark for isolated sign recognition, we only generate and assess labels derived from the 3D-LEX ASL vocabulary.

Our approach is designed to produce labels that resemble the handshape annotations typically found in ISR benchmarks, facilitating a meaningful comparison between automated and expert annotations. The glosses in ISR benchmarks are commonly assigned a single handshape label, based on the dominant handshape observed in a single reference video. We ensure that the number of possible label classes in our estimations corresponds approximately to the set of classes identified in the video-data benchmark WLASL. For the implementation and instructions on how to replicate our findings, please refer to the GitHub repository.

\paragraph{Temporal segmentation}
To differentiate characteristic handshape signals from any resting or transitional poses, we construct a temporal segmentation method by calculating the Euclidean distance to each frame relative to the calibration poses. This method enables us to perform a first-order discrimination of signals within a recording. 

\begin{figure}[!t]
    \centering
    \includegraphics[width = \columnwidth]{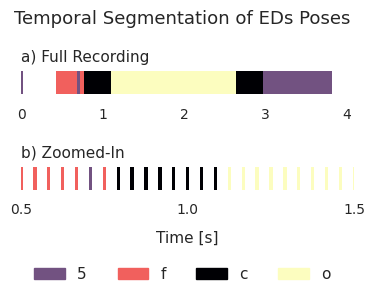}
    \caption{\textbf{Time-series visualization of handshape classification}: Classification of the ASL sign '\textit{zero}', labeled by experts with the handshape '\textit{o}'. Frames are captured and displayed as bars, and each bar's color indicates the handshape, determined by applying the Euclidean distance method frame-by-frame. White space indicates that no data was recorded at that time. The timeline, marked on the x-axis, spans four seconds for this sign. A detailed view at the 1-second mark is provided in the lower row for closer inspection. Our segmentation pipeline identifies the handshapes '\textit{5}', '\textit{f}', '\textit{c}', and '\textit{o}', selecting frames corresponding to '\textit{o}' as the characteristic signal of '\textit{zero}'.}
    \label{fig:ls}
\end{figure}

We estimate and segment the poses of both hands to take into consideration that the signer may not strictly enforce the use of their dominant hand. Subsequently, we calculate the frequency of each observed handshape and select the handshape with the highest frame count.  As the typically most frequent signal is the resting pose '\textit{5}', we only select the '\textit{5}' handshape if it is detected in more than \texttt{90}\% of the frames, otherwise, we select the second most frequently occurring class. The frames where the dominant handshape was detected are then selected as candidate frames for downstream analysis. Figure \ref{fig:ls} showcases the output of a Euclidean distance handshape classification approach on frames from the captured sign '\textit{zero}'. Here, the handshape '\textit{o}' was identified as the characteristic handshape of the sign.

\paragraph{Semi-automatic labeling}
The Euclidean distance labeling technique limits the identification of handshapes to those poses used during the glove calibration phase. This is suboptimal, as the calibration methodology of stretch sensors for capturing sign language is still in a nascent stage. Specifically, the calibration poses may not cover the full range of handshapes present in the lexicons.

To enable a more flexible identification of handshapes, we applied k-means clustering on the average poses of the frames selected during the temporal segmentation. We selected k=50, which is approximately the number of handshapes identified in ASL-LEX for the 3D-LEX vocabulary.  We assign a new handshape label to each sign in 3D-LEX ASL, corresponding to the arbitrary cluster IDs assigned while clustering the high-dimensional features.

\begin{figure}[!h]
    \centering
    \includegraphics[width = 0.9\columnwidth]{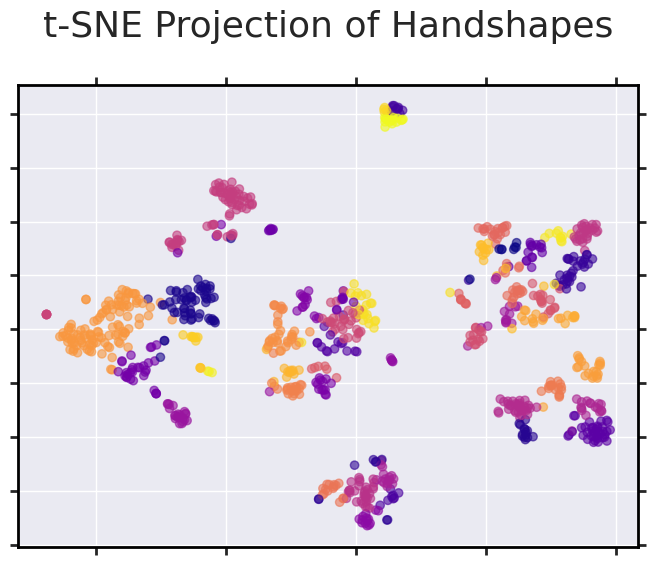}
    \caption{\textbf{t-SNE projection}: A t-SNE projection of average hand poses into two dimensions, where the poses were averaged across temporal segments of each sign determined by the Euclidean segmentation method. The projection space lacks units and aims solely to illustrate how high-dimensional 3D-LEX handshape features cluster, highlighting distinguishable signals. Each color represents one of 50 k-means cluster IDs, serving merely to aid visual differentiation of the clusters.}
    \label{fig:tsne}
\end{figure}

Figure \ref{fig:tsne} presents a t-SNE projection into two dimensions of the average hand poses, demonstrating that the high-dimensional features cluster. This implies that the signals from the gloves carry sufficient information to distinguish between different handshapes in sign language, revealing distinct characteristics for clusters of signs.


\paragraph{Evaluation of annotations}
To evaluate the efficacy of our annotations, we employed the OpenHands framework \cite{openhands2021}.  More precisely, we adopted the framework's adaptation as implemented by Kezar et al. \cite{kezar2023}, which facilitates gloss recognition supported by phonetic properties. Their foundational work demonstrated that training with phonetic labels enhances gloss recognition accuracy, by merging the WLASL benchmark with the expert linguistic descriptions provided by the ASL-LEX dataset. 

In our evaluation process, we trained an SL-GCN \cite{Jiang2021} architecture to predict glosses within the WLASL dataset, where we use the subset of the WLASL data which overlaps with the 3D-LEX vocabulary. Training persists until validation accuracy ceases to improve for 30 consecutive epochs. The distribution of files across the training, validation, and test splits utilized in our experiments is detailed in Table \ref{tab:split}.

To provide a baseline for comparison, we trained the SL-GCN to predict glosses both with and without leveraging handshape labels from the ASL-LEX. Subsequently, we substituted the ASL-LEX handshape labels with our semi-automatic annotations and retrained the models to undertake gloss recognition supported by our annotations. This approach facilitates a comparison of our semi-automatic annotation method against human expert annotations, in terms of their ability to support learning in a downstream ISR task. 

\begin{table}[!ht]
\begin{center}
\begin{tabularx}{\columnwidth}{|X|X|X|}
      \hline
        \textbf{Train} & \textbf{Val}& \textbf{Test} \\      \hline
        8209 &2174& 1668 \\ 
      \hline
\end{tabularx}
\caption{\label{tab:split} \textbf{Train-Val-Test splits:}
Number of examples in the Train-Val-Test splits for the WLASL benchmark experiments.}
 \end{center}
\end{table}

\paragraph{Results}

The outcome of our isolated sign recognition experiment using semi-automatic handshape labels is presented in Table \ref{tab:isrresults}. We provide the top-1 recognition accuracy on the test set, meaning the ratio of how often the model predicted the correct gloss as the most likely label for a video amongst 1,000 classes. As can be observed, the automatic annotations perform on par with annotations provided by linguistic experts. This is an indication that high-resolution 3D data can offer to reduce the costs associated with linguistic annotation of signs in video datasets and that StretchSense signals are adequate to capture essential handshape features in signs. 

\begin{table}[!ht]
\begin{center}
\begin{tabularx}{\columnwidth}{|X|X|X|}
      \hline
         $a_1^{\textnormal{N}}$ &$a_1^{\textnormal{E}}$&$a_1^{\textnormal{A}}$\\\hline
        $0.44_{\pm0.01}$&$0.48_{\pm0.01}$&$\mathbf{0.49_{\pm0.01}}$\\
      \hline
\end{tabularx}
\caption{\label{tab:isrresults} \textbf{Top-1 recognition accuracy}: Accuracy using no (N) handshape labels, expert (E) labels and automatic (A) labels. The accuracies are averaged across 8 runs, and the standard deviation across measurements is provided in the subscripts. }
 \end{center}
\end{table}


\section{Limitations and Prospects}\label{sec:lim}
In the process of capturing our data, we have observed many potential areas for improvement. In this section, we highlight some of the current limitations in our methodology, and our intent for addressing them in future work. 

Like numerous datasets in sign language research, a significant limitation of 3D-LEX is signer diversity. A dataset comprising a single example for each sign, and which contains only five signers, is insufficient for representing the diversity and rich prosody inherent to sign languages. It is as such not possible to use 3D-LEX in isolation to learn representations useful in sign applications. Consequently, 3D-LEX can primarily serve for limited feature studies or to support video datasets by either providing a 3D ground truth or synthesizing multi-view 2D data from one signer. Future work should consider exploring 3D data which includes both multiple examples per signer and multiple signers per gloss. 

While all participants were native signers, it is critical to highlight that only one had ASL as their primary language. As a result, a significant segment of the 3D-LEX ASL dataset was produced by signers whose primary language is NGT  but who were proficient in ASL. The impact of employing signers whose primary sign language differs from the captured target language, on the quality and authenticity of lexical sign data remains an area for future research. This concern is recognized as a limitation in v1.0 of 3D-LEX.

The dataset has a limited scope, which comprises a non-exhaustive set of phonological features and vocabularies from the complete languages. However, our method facilitates the production of larger vocabularies and data for additional sign languages.

We observed several limitations in our current pipeline. While experimenting with the data acquisition control we noticed varying preferences among signers for operating the pedal. The choice of operator resulted in the emergence of several distinct patterns within the data. When signers themselves operate the pedal, it's generally more efficient but introduces a signal from foot movement at the start and end of each sign. Conversely, using an external operator can result in greater variability in the timing of recordings, affecting the consistency of the recorded time window around each sign. Efforts to streamline these production elements are anticipated in future work.

While our system has been designed with a focus on efficiency, we have identified several limitations concerning the hardware. To the best of our knowledge, 3D-LEX is the first publicly available dataset using the StretchSense gloves to conduct statistical analysis on handshapes in sign language. These gloves were initially developed to generate animation data, which typically does not require the same degree of accuracy in capturing detailed, varied and intricate movements of fingers and hands.  Therefore, employing these gloves to provide detailed studies of handshapes in sign language represents a novel and experimental approach. Although the gloves have shown promising capabilities, their performance has presented several challenges. 

Notably, the precision of the gloves' measurements is closely tied to how well they fit the signers' hands and the length of time they are worn. A snugger fit typically leads to higher accuracy. However, prolonged usage has been observed to decrease accuracy, likely due to the glove's position shifting on the hand, thereby deviating from its calibrated stance. Shifts can occur for example when hands swell from accumulated heat and from natural movements during wear. Larger gloves relative to the hand size are more prone to positional shifts, exacerbating this issue.

\begin{figure}[!t]
    \centering
    \begin{subfigure}{0.23\columnwidth}
        \includegraphics[height=2cm]{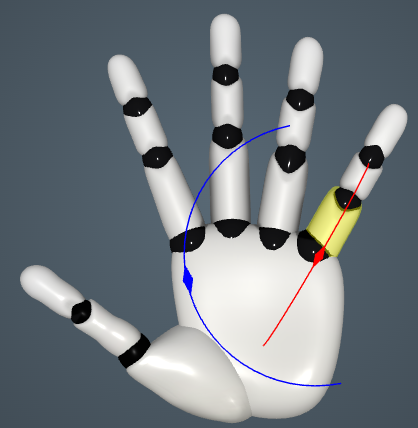}
        \caption{'\textit{five}'}
        \label{fig:sub1}
    \end{subfigure}
    \hfill 
    \begin{subfigure}{0.23\columnwidth}
        \includegraphics[height=2cm]{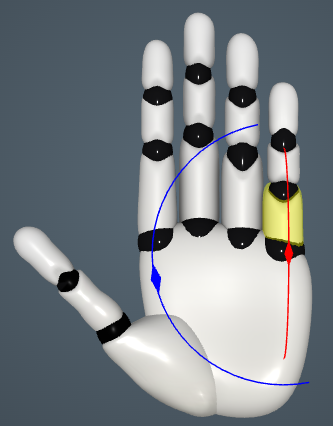}
        \caption{'\textit{open b}'}
        \label{fig:sub2}
    \end{subfigure}
    \hfill
    \begin{subfigure}{0.11\columnwidth}
        \includegraphics[height=2cm]{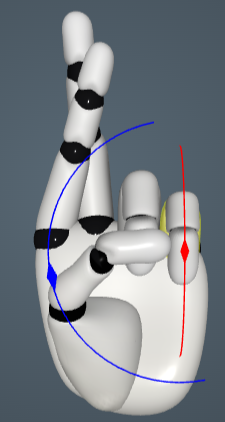}
        \caption{'\textit{r}'}
        \label{fig:sub3}
    \end{subfigure}
    \hfill
    \begin{subfigure}{0.11\columnwidth}
        \includegraphics[height=2cm]{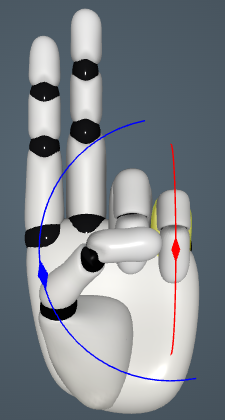}
        \caption{'\textit{h}'}
        \label{fig:sub4}
    \end{subfigure}
    \hfill
    \begin{subfigure}{0.19\columnwidth}
        \includegraphics[height=2cm]{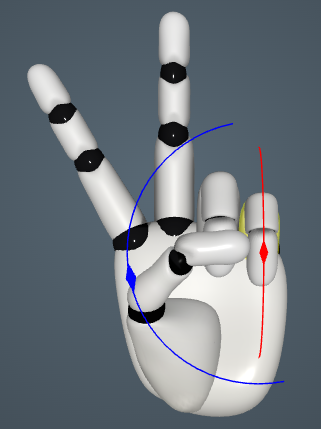}
        \caption{'\textit{v}'}
        \label{fig:sub5}
    \end{subfigure}
    \caption{\textbf{Failure modes using the StretchSense gloves}: Example handshapes that are challenging to discern for the gloves, conditioned on our calibration scheme. The gloves struggled to differentiate between the handshapes '\textit{five}' and '\textit{open b}', and between the handshapes '\textit{r}', '\textit{h}' and '\textit{v}'.}
    \label{fig:failure_mode}
\end{figure}

The Hand Engine software is prone to overfitting the sensor data to the calibration poses, a tendency that amplifies when training involves an extensive calibration pose set. Currently, the calibration process utilizes either 20 or 25 poses. We observed that such a detailed pose repertoire complicates Hand Engine's ability to accurately replicate more complex poses and distinguish between poses where the shift in stretching values are relatively small. Figure \ref{fig:failure_mode} illustrates a series of poses that exhibit substantial differentiation challenges for the gloves under our calibration framework. With the current version of Hand Engine, future research may gain advantages from employing a smaller set of calibration poses. Ideally, these selected handshapes should not only be representative of those within the dataset but also exhibit maximum distinction from each other within the calibration set.

An in-depth assessment of calibration methods to address overfitting issues warrants further exploration. This becomes especially critical in capturing continuous signing, where the range of anticipated handshapes is far more variable and unpredictable than in lexical datasets. The 3D-LEX team is actively engaging with StretchSense to enhance glove calibration for sign languages, focusing on better support for continuous signing and capturing a broader spectrum of handshapes. The gloves' ability to accurately represent signing is contingent upon the calibration process, however, as this is a software concern, we expect the conditions for continuous signing to improve in later versions of the Hand Engine software. 

Upon assessing the Vicon data, we identified several artifacts occasionally occurring in recordings. For example, we observed random hand orientation flips, which can be attributed to occlusions, where the cameras lost clear line-of-sight to the hands. In such instances the markers may be mistaken for each other, causing the palm to rotate when displayed on an avatar. To mitigate this issue, one can attempt to optimize the positions of the cameras standing on the floor or apply post-processing techniques, such as the filter and gap solver functionalities available in Shogun Post, or by re-labeling the swapped markers. 

Moreover, due to limited time, we could not assess the data generated by the LLF application in detail. However, we observed considerable variation in the use of markers like mouthing cues and gaze among participants. In our future research, we aim to delve into these patterns and thoroughly evaluate the quality of the facial feature data.

In our evaluations of 3D-LEX, we presented a basic approach to deriving annotations. However, we emphasize that signs are complex and may contain transitions or oscillate between multiple characteristic handshapes throughout the execution of a sign. While our method approximates the dominant handshape, there are potential benefits in deriving more sophisticated annotation strategies, which consider these transitions and oscillations, and potentially provide multiple phonetic properties for the handshape per sign. However, it is noteworthy that, even in the nascent stages of developing the 3D-LEX production methodology, our automatic annotations yield benefits comparable to those derived from leveraging annotations provided by experts.

\section{Privacy and Ethical Considerations} \label{sec:eth}
The success of machine learning methods has led to large increases in requests for data. While this implies heightened concerns for privacy across computational sciences, it is important to recognize that data collection from minority language communities is at particular risk: Both because a status as deaf classifies as sensitive information, but also because data collection from small populations limits anonymity \cite{Bragg2020}. Additionally, certain sign language datasets that are publicly accessible were compiled without obtaining informed consent from the individuals featured, particularly those datasets that gather information from platforms such as YouTube. All signers contributing to the production of 3D-LEX gave informed consent and received compensation. Moreover, the anonymity of contributors is enhanced compared to typical video datasets, since the motion capture recordings do not visually reveal the signers. To further protect signer anonymity, each participant has been assigned a unique signer ID.

\section{Conclusion}
In this paper, we introduce a new and efficient method for collecting 3D sign language data, resulting in the 3D-LEX dataset, and describe a semi-automatic approach for producing phonetic annotations. The 3D-LEX dataset was produced leveraging three distinct motion capture systems, with two collection techniques to capture manual markers and one technique to capture non-manual markers. Although our approach shows considerable room for improvement, we highlight its potential by automatically generating handshape labels for 1,000 ASL signs. Our initial evaluations of the labels on a downstream ISR task reveal that the semi-automatic annotations offer benefits parallel to those of expert annotations. In conclusion, the 3D-LEX v1.0 demonstrates considerable potential even in its early stages of development. We anticipate that future research using 3D-LEX will investigate synthesizing multi-view data from the 3D ground truths to support tasks such as multi-view SLR, and develop approaches annotating additional phonetic classes.

\section{Acknowledgments}
The team behind 3D-LEX consisted of both Deaf and hearing researchers, whose participation in the project was made possible through financial support from the Platform Digital Infrastructure for the Social Sciences and the Humanities (PDI-SSH) and the Netherlands Organization for Scientific Research (NWO). We extend our gratitude to all external participants who assisted in gathering the data, and to the \href{https://visualisationlab.science.uva.nl/}{Visualisation Lab} and \href{https://www.signlab-amsterdam.nl/}{SignLab} at the University of Amsterdam for generously providing us with their facilities and equipment. In addition, we wish to thank the company ProCare\footnote{\href{https://www.procarebv.nl/}{ProCare BV, The Netherlands}}, who performed the setup of our Vicon rig.


\section{Bibliographical References}\label{sec:reference}

\bibliographystyle{lrec-coling2024-natbib}
\bibliography{lrec-coling2024-example}

\section{Language Resource References}
\label{lr:ref}
\bibliographystylelanguageresource{lrec-coling2024-natbib}
\bibliographylanguageresource{languageresource}

\end{document}